\documentclass[letterpaper, 10 pt, conference]{ieeeconf}  

\IEEEoverridecommandlockouts                              

\usepackage{booktabs}
\usepackage{multirow}
\usepackage{graphicx}
\usepackage{amsmath}
\usepackage{amssymb}
\usepackage{xcolor}
\usepackage{hyperref}
\usepackage{booktabs}
\usepackage{tabularx}
\usepackage{pifont}
\usepackage{comment}
\usepackage{censor}
\usepackage{siunitx}
\usepackage[normalem]{ulem}

\hypersetup{
    colorlinks=true,  
    linkcolor=black,  
    citecolor=black,  
    filecolor=black,  
    urlcolor=blue     
}

\newcommand{\normtotex}{Norm2Tex}

\title{\LARGE \bf Norm2Tex: Augmenting Visuo-Tactile Simulations with Texture}
\StopCensoring 
\author{\censor{Seongjin Bien$^{*1}$, D\'ebora Oliveira Makowski$^{*1}$, Roberto Calandra$^{2}$, Florian Walter$^{3}$ and Wolfram Burgard$^{1}$}
\thanks{\censor{$^*$Equal contribution.}}%
\thanks{\blackout{$^{1}$University of Technology Nuremberg, Germany. Contact email: \texttt{seongjin.bien@utn.de}.}}%
\thanks{\blackout{$^{2}$Technical University of Dresden, Germany.}}
\thanks{\blackout{$^{3}$Deggendorf Institute of Technology, Germany.}}
\thanks{Available at \url{https://utn-air.github.io/norm2tex}
}
}

\begin{document}
\bstctlcite{IEEEexample:BSTcontrol}

\maketitle
\thispagestyle{empty}
\pagestyle{empty}

\begin{abstract}
Large-scale datasets are essential for training generalist robot control policies. 
Collecting real-world tactile data is costly and time-consuming, motivating the use of tactile simulations.
However, current tactile simulators capture only overall contact geometry and miss fine details like texture. 
This results in a significant domain shift between simulated and real tactile data.
To address this gap, we introduce \normtotex, a plug-in method that augments simulations of vision-based tactile sensors with high-frequency surface details from normal map textures. 
By modifying the target object’s depth map before a tactile simulator’s rendering pipeline, \normtotex\ seamlessly integrates into different tactile simulators. 
We also evaluate sim-to-real transfer using material classification and a reinforcement learning task. 
Our results show that \normtotex\ preserves material-dependent tactile information across domains, improving texture recognition and producing material-dependent control behavior in the real world. 

\end{abstract}

\section{Introduction}
Vision-based tactile sensors ~\cite{digit,gelsight, wang2021wedge,liu2026crossmodalvisuotactilerepresentationlearning} have enabled high-density tactile sensing by converting contact-induced surface deformations into RGB images. They typically use a miniature camera positioned behind a deformable contact surface, such as silicone, to capture its deformation upon contact. The resulting images encode information across multiple spatial scales: low-frequency deformations capture the overall geometry of the contacting object, while high-frequency variations reveal fine surface details, including texture. This ability to observe micrometer-level details has motivated the development of applications such as slippage detection~\cite{improvedgelsight}, texture classification~\cite{whatmatters2024,fabricsorting2026}, and material defect identification~\cite{fabricdefect2022,defectdetection2026} 

Despite their rich sensing capabilities, a major challenge in applying vision-based tactile sensors to robotic manipulation is the collection of large-scale tactile datasets. This is particularly limiting for deep learning-based methods, whose performance typically depends on access to a substantial amount of training data. Unlike third-person perspective cameras, which are common in many setups and can continuously capture meaningful information throughout a task episode, tactile sensors provide information only once physical contact is established. Consequently, collecting diverse tactile observations requires repeated contact-rich physical interactions, making data acquisition considerably more costly.

\begin{figure}[t]
  \vspace{5.1pt}
  \centering
    \includegraphics[width=\linewidth]{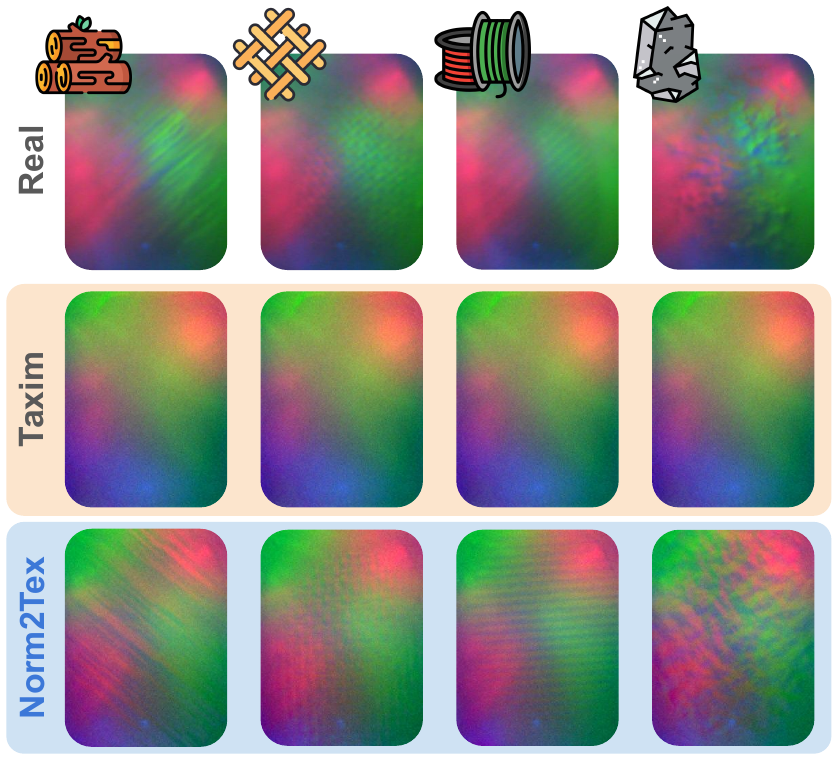}
        \caption{Comparison of real-world visuo-tactile data from DIGIT (top) and the commonly simulated rendering on Taxim (center). Independent of the texture, the Taxim rendering looks the same.  \normtotex\ (bottom) fills this sim-to-real gap by augmenting surface details (left to right: wood, fabric, 3D print, and rock) to the rendering.}
        \label{fig:eyecandy}
\vspace{-1.5em}
\end{figure}

Several tactile simulators have been developed to address this problem~\cite{Wang2022TACTO,si_taxim_2021,tacsl_2025,tacflex_2025}. However, visuo-tactile datasets collected in simulation suffer from a substantial sim-to-real gap. To avoid the inherent difficulties of modeling the dynamic properties of compliant contact surfaces and their interactions with other objects, Current methods typically assume prior knowledge of the contacting object’s geometry, commonly represented by or derived from a mesh, and use this information to compute the contact geometry required for tactile deformation and rendering~\cite{difftactile_2024,tacipc_2024,tacchi_2023,tacchi2_2026}. As a result, high-frequency surface properties, such as textures and fine geometric details, can only appear in the simulated tactile image if they are explicitly represented in the object's source geometry. Modeling such details at the mesh level is costly and cumbersome, and adds additional complexity when calculating contact. As shown in Fig.~\ref{fig:eyecandy}, the lack of detailed object geometry limits the sim-to-real transferability of existing tactile simulation approaches because real-world objects exhibit a wide variety of surface textures and fine-scale structures. Such variations are ubiquitous in everyday environments and can substantially affect the tactile observations produced during contact. It remains difficult to replicate them efficiently with existing simulation pipelines.

In this work, we introduce \normtotex, a novel method that addresses this limitation in the current landscape of tactile simulation. Rather than serving as a standalone simulator, \normtotex\ is a drop-in extension for existing tactile simulators that use a depth-map representation as in TACTO~\cite{Wang2022TACTO}, Taxim~\cite{si_taxim_2021} and Tacchi~\cite{tacchi_2023}. Given a target object's normal map, \normtotex\ modifies this intermediate representation to inject high-frequency surface details, then passes the result through the remainder of the simulator's original rendering pipeline. In this way, \normtotex\ enables existing simulators to represent fine-scale texture information without requiring these details to be explicitly encoded in the object's underlying mesh geometry.

Our contributions are fourfold. First, we introduce \normtotex, a lightweight drop-in extension for existing depth map-based tactile simulators that enables rendering high-frequency texture details at low computational overhead. Second, we provide an extensible Blender-based~\cite{blender} procedural texture generation pipeline capable of producing normal maps for arbitrary meshes. It covers four classes of common textures with a wide range of configurable parameters. Third, we evaluate \normtotex's efficiency by integrating it into two existing tactile simulators, TACTO and Taxim, and assess its zero-shot sim-to-real transfer capabilities on both a texture classification task and a reinforcement learning-based grasp force adjustment task. Finally, we open-source \normtotex\ and all related materials.

\section{Related Work}
\subsection{Simulators for Vision-Based Tactile Sensors}
Existing tactile simulators differ primarily in how they obtain and represent the deformation of the sensor's compliant surface. Several approaches approximate the contact geometry directly from depth observations. The GelSight simulator~\cite{gelsight} captures a depth image at the contact region and processes it to obtain a surface representation that is subsequently rendered using a Phong illumination model. Taxim~\cite{si_taxim_2021} similarly operates on contact depth information, but uses calibrated, data-driven mappings between local surface geometry and tactile pixel intensities to synthesize the resulting image. TacSL~\cite{tacsl_2025} captures depth observations from a virtual tactile camera and maps the resulting surface geometry to RGB values through a calibrated optical model. TACTO~\cite{Wang2022TACTO} also supports depth-based rendering, including an interface that converts an input depth map into a mesh representation of the deformed gel surface.

More recent simulators increasingly incorporate explicit physical models of elastomer deformation before tactile image rendering. TacIPC~\cite{tacipc_2024}, TacFlex~\cite{tacflex_2025}, and DiffTactile~\cite{difftactile_2024} employ finite element methods (FEM) to model the compliant sensor surface, while the Tacchi family~\cite{tacchi_2023, tacchi2_2026} uses the Material Point Method (MPM) to simulate elastomer deformation. These approaches can more accurately capture contact mechanics and provide additional physical information, such as marker displacement induced by deformation. The resulting deformed surface is then converted into a representation suitable for optical rendering. For example, Tacchi reconstructs a continuous depth map from its simulated surface particles before generating the tactile image.

\normtotex\ is complementary to these approaches because it operates at the interface between the simulated contact surface and the subsequent tactile rendering stage. For simulators that expose the deformed contact surface as a depth-map representation, \normtotex\ can augment this intermediate representation with high-frequency surface details without modifying the underlying contact or elastomer simulation. This allows improvements in fine-scale tactile appearance to be incorporated independently of how the underlying deformation is obtained.

\subsection{Textures in Simulated Tactile Images}

Existing works have proposed learning-based methods to address the lack of high-frequency surface details and other appearance discrepancies in simulated tactile images. Jianu \emph{et al.}~\cite{jianu_reducing_2022} introduced a texture generation network that transforms simulated tactile images into more realistic observations by learning textural artifacts from real tactile data and applying them specifically within the contact region. Subsequent work employed a bidirectional CycleGAN~\cite{CycleGAN2017,texture_cyclegan_2022} framework to translate between simulated and real GelSight image domains, primarily learning the non-ideal optical characteristics that are difficult to reproduce analytically~\cite{chen_bidirectional_2022}. SightGAN~\cite{azulay_augmenting_2024} builds on CycleGAN-style image translation by operating on tactile difference images and adding constraints to better preserve contact patterns and embedded force information while generating real-like observations~\cite{azulay_augmenting_2024}. Others have explored diffusion-based generative models for synthesizing realistic tactile observations. Tactile Diffusion uses a conditional diffusion model to directly generate tactile images from simulated contact depth, allowing complex illumination and appearance variations to be learned from real sensor observations~\cite{higuera_perceiving_2023}. More recently, Lin \emph{et al.}~\cite{lin_diffusion_2025}~proposed a contact condition-guided diffusion model that conditions tactile image generation on object RGB images together with contact-force information, enabling the synthesis of high-fidelity tactile observations under varying contact conditions and demonstrating the reconstruction of fine-scale surface textures.

Although these deep learning-based approaches can generate high-quality tactile appearances and textural artifacts, adapting them to different sensor characteristics or types of surface variation typically requires collecting real tactile data for training or fine-tuning. In contrast, \normtotex\ can directly leverage either procedurally generated normal maps or existing normal maps without requiring additional tactile observations. 

Finally, image generation approaches do not inherently guarantee that the generated fine-scale appearance remains spatially consistent across consecutive observations. This is particularly important for vision-based tactile sensing, where the temporal evolution of contact provides information about how an object moves relative to the sensor during manipulation. In \normtotex, surface textures are explicitly defined by a UV-mapped normal map associated with the target mesh, ensuring that fine-scale surface details remain spatially anchored to the object geometry. Consequently, observations of the same physical surface region produce consistent high-frequency tactile features across successive frames, while those features evolve coherently as the contact location changes over time.

\section{Method}

\begin{figure*}[t]
  \centering
  \vspace{5.1pt}
  \includegraphics[width=\textwidth]{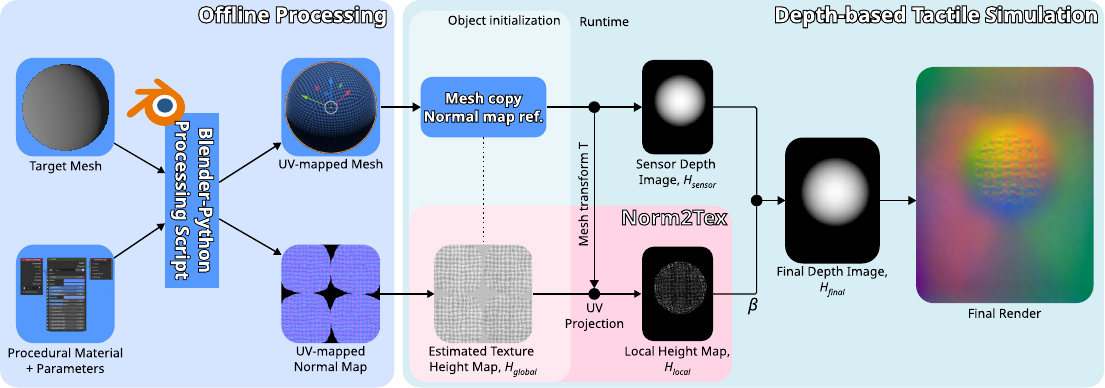}
  \caption{
    The \normtotex\ architecture. First, a procedural texture is created for a given mesh using Blender. At initialization, the global texture height map $H_\text{global}$ is computed, and a copy of the mesh is created internally for reference. During simulation, given the relative transform of the mesh to the sensor at the contact point, the corresponding UV coordinates are used to fetch the pixels from $H_\text{global}$ to create $H_\text{local}$. Finally, $H_\text{sensor}$ and $H_\text{local}$ are added together to create the final depth image $H_\text{final}$, which is then used for rendering the tactile image by the tactile simulator. }
  \label{fig:meth_arch}
  \vspace{-1em}
\end{figure*}

\subsection{Preliminaries}\label{meth:prelim}
A normal map is first calculated using a mesh's original, high-resolution vertices. This map is then applied to a simplified, lower-resolution version of the mesh~\cite{Cohen1998AppearancepreservingS}. This makes the simpler mesh look more detailed by mimicking how a detailed mesh reflects light. This process saves computational resources during runtime. A UV normal map represents the tangent-space unit normals of the corresponding mesh. For each texture coordinate $(x, y)$, it stores a tangent-space unit normal in its RGB values as:
\begin{equation}
    \mathbf{c}(x, y) = (c_x, c_y, c_z) \in [0,1]^3
\end{equation} 
Remapping the RGB channels from [0, 1] to [-1, 1] and normalizing them allows us to recover the signed normal
\begin{equation}
\widetilde{\mathbf{n}}(x,y) = 2\mathbf{c}(x,y)-1, 
\end{equation}
\begin{equation}
\mathbf{n}(x,y) = \frac{\widetilde{\mathbf{n}}(x,y)}{\left\|\widetilde{\mathbf{n}}(x,y)\right\|} = (n_x,n_y,n_z) \in [-1, 1]^3.
\end{equation}

This value is proportional to the cross product of the surface tangent vectors in the x and y directions:
\begin{equation}
    t_x = (1, 0, h_x), t_y=(0,1,h_y),
\end{equation}
\begin{equation}
    t_x \times t_y = (-\frac{\partial h}{\partial x}, -\frac{\partial h}{\partial y}, 1) \propto \mathbf{n},
\end{equation}
where $h_x$ and $h_y$ are the true, unrecoverable height that created the normal map. This allows us to estimate the $x$- and $y$-direction gradients:
\begin{equation}
d_x = -\frac{n_x}{n_z}, \qquad d_y = -\frac{n_y}{n_z}.
\end{equation}

\subsection{Generating Textures for Arbitrary Meshes}
\autoref{fig:meth_arch} shows an overview flow diagram of \normtotex. Our proposed pipeline requires that a given mesh comes with its corresponding UV-matched normal map. However, freely available 3D model collections typically include only RGB textures, not normals. For vision-based tactile sensor simulation, the goal is not to use normal maps from higher-resolution meshes. Instead, we want to capture the fine textural details of the material, like wood grain, that the mesh is meant to represent. This detail lets us design spatially consistent procedural materials in Blender that can generate normal maps for arbitrary meshes. 

\begin{figure}[t]
  \vspace{5.1pt}
  \centering
    \includegraphics[width=0.23\linewidth]{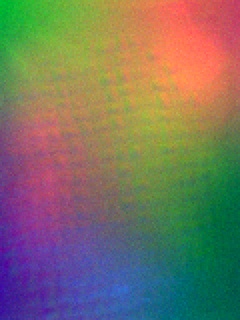}
    \includegraphics[width=0.23\linewidth]{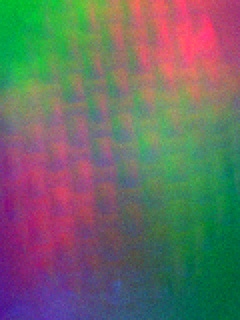}
    \includegraphics[width=0.23\linewidth]{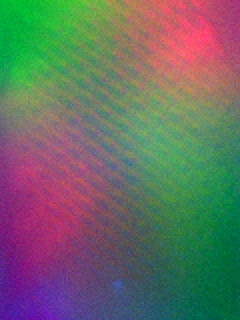}
    \includegraphics[width=0.23\linewidth]{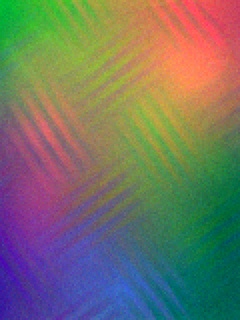}
        \caption{Examples of different fabric renderings generated from a single procedural texture (\textit{fabric}) by adjusting its parameters. }
        \label{fig:fabric_showcase}
\end{figure}

Such procedural materials are usually created by combining various noise and structural components to mimic the desired real-life material, each with its own adjustable parameters. This allows a single base material to be customized as needed, including its texture resolution. \autoref{fig:fabric_showcase} shows several images generated from a single {fabric} material. We designed several such procedural materials corresponding to commonly found surface types. Additionally, we implemented a Python-based script to easily apply these materials and their parameters to arbitrary meshes to generate the RGB texture, the normal map, and a UV-adjusted copy of the input mesh, which we then use for the rest of the pipeline.

\subsection{Converting RGB Normal to Height Map}

After generating the normal map, we convert it into a pseudo-height map using Fourier-domain Poisson integration based on the integrability projection of Frankot and Chellappa~\cite{frankot1988}. The method projects the surface-gradient field derived from the normal map onto the space of integrable gradient fields in the Fourier domain, yielding a least-squares estimate of the corresponding surface height. 

Because the normal map consists of independent UV islands corresponding to different mesh surface regions, we first divide it into $N$ image patches. For each patch $p\in N$, we decode the normal-map pixels into unit normals and calculate the associated gradient field
\[\mathbf{d}_p(x,y) = 
\begin{bmatrix}
    d_{p,x}(x,y) \\
    d_{p,y}(x,y)
\end{bmatrix},\]
using the method summarized in \autoref{meth:prelim}.

Let $h_{p,i,j}$ denote the unknown pseudo-height at pixel $(i,j)$ belonging to patch $p$. We seek a height map $H_p$ whose horizontal and vertical finite differences most closely match the normal-derived gradients. This can be expressed as the least-squares problem
\begin{equation*}
\begin{aligned}
E(H) = \sum_{i,j} \Big[
&\left(D_x(h_p)-d_{p,x}(i,j)\right)^2 \\
+{}&
\left(D_y(h_p)-d_{p,y}(i,j)\right)^2
\Big]
\end{aligned}
\end{equation*}
where $D_{x}(h) =  h_{i+1,j}-h_{i,j}$ and $D_{y}(h) =  h_{i,j+1}-h_{i,j}$ are the neighbor difference operator.
The normal-derived gradient field need not be exactly integrable because of quantization, filtering, or inconsistencies in the normal map. Consequently, the objective finds the height field whose discrete gradient is closest to $\mathbf{d}_p$ in the least-squares sense.

Setting the derivative of $E$ with respect to every $h_{i,j}$ to zero produces the discrete Poisson equation
\[
\Delta_d H_p
=
\nabla_d \cdot \mathbf{d}_p,
\]
where $\Delta_d$ is the discrete five-point Laplacian and $\nabla_d \cdot \mathbf{d}_p$ is the discrete divergence of the target gradient field.

Solving this equation directly would require solving a large linear system. Under periodic boundary conditions, however, the discrete Fourier basis diagonalizes the Laplacian, allowing the solution to be computed efficiently using the discrete Fourier transform. First, the divergence is transformed into the frequency domain:
\[
\widehat{D}(u,v)
=
\mathcal{F}
\left\{
\nabla_d \cdot \mathbf{d}_p
\right\}(u,v).
\]
For a patch of width $W_p$ and height $H_p$, the eigenvalue of the periodic discrete five-point Laplacian at frequency $(u,v)$ is
\[
\lambda_p(u,v)
=
2\cos\left(\frac{2\pi u}{W_p}\right)
+
2\cos\left(\frac{2\pi v}{H_p}\right)
-4.
\]
The frequency-domain height map is therefore obtained as
\[
\widehat{H_p}(u,v)
=
\frac{\widehat{D}_p(u,v)}{\lambda_p(u,v)},
\qquad
(u,v)\neq(0,0).
\]

At $(u,v)=(0,0)$, the Laplacian eigenvalue is zero. This reflects the fact that the gradient field determines the reconstructed height only up to an arbitrary additive constant. We therefore set
\[
\widehat{H}_p(0,0)=0,
\]
which selects the zero-mean solution. Finally, applying the inverse Fourier transform yields the reconstructed pseudo-height map:
\[
H_p
=
\mathcal{F}^{-1}
\left\{
\widehat{H}_p
\right\}.
\]
This process is repeated for all $N$ UV island patches, and the processed patches $[H_1,...H_N]$ are then composed together back into their original positions in the input image space to preserve their UV coordinate mapping.

We perform this reconstruction process twice. Before each reconstruction, we filter the decoded normal map with a mask-aware Gaussian filter and renormalize it. A fine pass with $\sigma_\text{fine}=0.35$ preserves local details, while a coarser pass with $\sigma_\text{coarse}=1.1$ suppresses high-frequency variations. The resulting height maps $H_{fine}$ and $H_\text{coarse}$ are then combined with a weight $\alpha=0.2$ and normalized using the combined height map's 0.05th and 99.95th percentiles:
\[H_\text{global} = \text{normalize}\big((1-\alpha)H_\text{fine} + \alpha H_\text{coarse}\big)\]
We implement three types of normalization: [0, -1], [-1, 1] and [0, 1], reflecting whether the texture adds or removes material from the underlying geometry, which can be configured at initialization.

\subsection{Querying Height Map Information During Simulation}
$H_\text{global}$ retains the original UV mapping of the input normal map. Since tactile simulators already require the relative pose of the object to the sensor for calculating the base height map $H_\text{sensor}$, we can reuse this information to calculate relevant pixels from $H_\text{global}$ which correspond to the surface of the mesh that is in contact with the sensor at simulation time.

Given the final rendered tactile image's resolution and an estimate of the millimeter-per-pixel value that the image represents, we query the height values using the in-contact mesh faces' UV coordinates and project them onto $H_\text{global}$ to retrieve the relevant pixels. This results in a local height map, $H_\text{local}$, with identical dimensions and valid pixels to $H_\text{sensor}$. Then, we multiply $H_\text{local}$ by a configurable value $\beta$, typically corresponding to how ``bumpy'' the given material is (e.g., 0.5 mm), and add the two height maps together to obtain the final height map as
\[H_\text{final} = H_\text{sensor} + \beta H_\text{local}.\]
This variable is then passed to the tactile simulator's original pipeline to complete the render.

\section{Experiments}
We perform three separate experiments to validate the usefulness of \normtotex. The first experiment assesses its applicability to two existing tactile simulators, TACTO and Taxim, and how it impacts their performance. The second experiment measures sim-to-real capability through a zero-shot classification task. The final experiment similarly assesses the sim-to-real capability of a tactile policy trained on \normtotex\ data. 

\subsection{Applicability and Performance Impact}
\begin{table}[t]
\vspace{5.1pt}
\centering
\caption{Comparison of render time for TACTO and Taxim with and without \normtotex, averaged over 1000 frames.}
\label{tab:timing}
\begin{tabular}{lcc}
\toprule
Condition & Total (s) & Difference  (\%) \\
\midrule
Taxim (CPU)             & 0.1019 & - \\
Taxim +\normtotex\ (CPU)  & 0.1106 & 8.54\% \\
Taxim (CUDA)            & 0.0130 & -\\
Taxim +\normtotex\ (CUDA) & 0.0132 & 1.54\% \\
TACTO (CPU)             & 0.3639 & -\\
TACTO +\normtotex\ (CPU)  & 0.3701 & 1.70\% \\
\bottomrule
\end{tabular}
\end{table}

To assess the performance of \normtotex, we implemented our method for two popular tactile simulations, Taxim and TACTO. For TACTO, we used its depth-based rendering pipeline, as the regular method does not require a depth map. We set up an equivalent MuJoCo environment~\cite{mujoco} consisting of a box mesh with 8 vertices and a corresponding fabric texture, and calculated the average rendering time over 1000 frames. The timings are listed in \autoref{tab:timing}. 

These results show that \normtotex\ has only a minimal impact on the overall performance of the respective simulators. The larger difference in Taxim is primarily due to its shadow post-processing, which requires more compute as the input depth map becomes more complex. With CUDA, the difference drops to 1.5\%, indicating that \normtotex's texture lookup is amenable to GPU-based computing.

\subsection{Zero-shot Sim-to-Real Material Classification}

\begin{figure}[t]
  \vspace{5.1pt}
  \centering
    \includegraphics[width=0.49\linewidth]{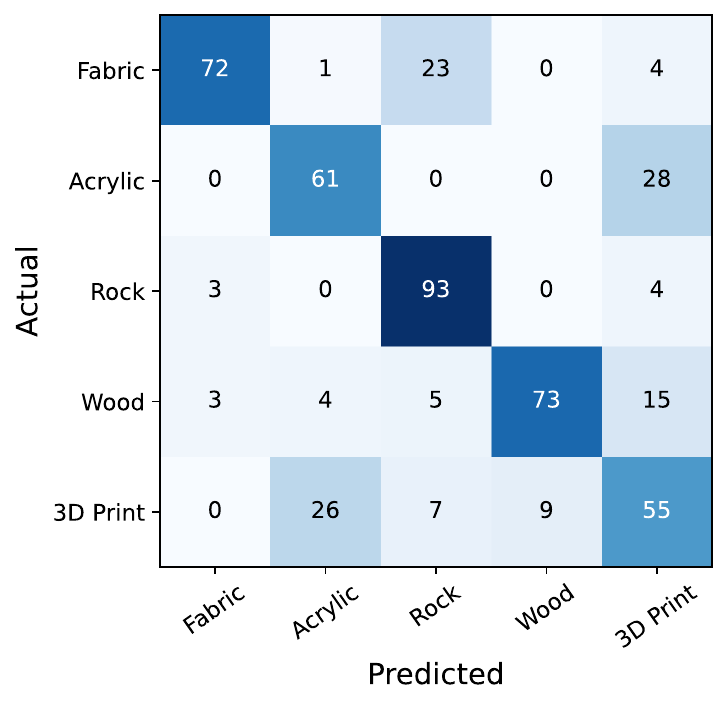}
    \includegraphics[width=0.49\linewidth]{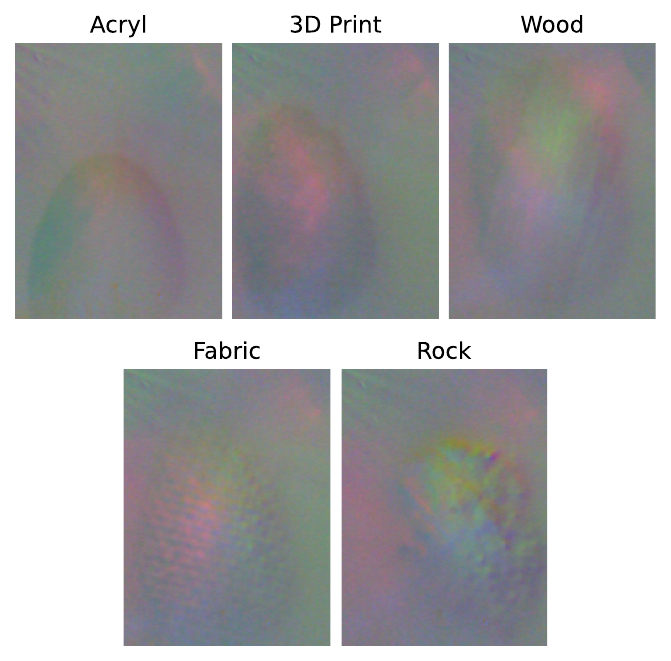}
        \caption{(Left) Confusion matrix for the classifier trained on simulated \normtotex\ data and deployed zero-shot in the real world. (Right) Real-world input image to the classifier, i.e., DIGIT image minus initial blank image.  Note that acrylic and 3D-printed smooth textures are hard to distinguish.
        When the fabric texture is granular, it can be mistaken for rock.}
        \label{fig:confusion}
\end{figure}

\begin{table}[t]
\vspace{5.1pt}
\centering
\caption{Per-material classification accuracy and precision.}
\label{tab:material-accuracy-precision}
\begin{tabular}{lcc}
\toprule
Material & Accuracy (\%) & Precision (\%) \\
\midrule
Fabric   & 72.0 & 92.3 \\
Acrylic  & 68.5 & 66.3 \\
Rock     & 93.0 & 72.7 \\
Wood     & 73.0 & 89.0 \\
3D Print & 56.7 & 51.9 \\
\midrule
Overall  & 72.8 & 74.4 \\
\bottomrule
\end{tabular}
\end{table}

To assess whether textures generated by \normtotex\ can assist sim-to-real transfer, we designed a classification task across different materials. We collected 100 real-world data samples for each of the five material categories (fabric, rock, smooth, 3D print, and wood) using a DIGIT sensor. We collected fabric samples from upholstered chairs and table mats. We sampled rock data from common pebbles, smooth data from porcelain plates and plastic water bottles, 3D print data from a 3D-printed cube, and wood data from a wooden plank purchased at a hardware store. 

Using the real data, we first manually tuned the Blender material parameters to match their appearances, then ablated on them with minor adjustments to increase the diversity of the generated texture. We then used the Taxim simulator with a MuJoCo-based data-collection script to create contact between a simulated sensor and a given geometry at randomly sampled locations. After contact, we performed small local motions to simulate different contact points on the DIGIT sensor's silicon pad. We collected 6000 Taxim renderings for each category. 

We trained a ResNet-18~\cite{he2016deep} model with a small classifier head on a cross-entropy objective using only simulated data, and tested it on real data in a zero-shot fashion. The classifier head is a single linear projection layer with dropout regularization. We initialized the network with ImageNet weights and freezed its weights up to the 8th residual block. To facilitate sensor invariance, we subtracted the neutral no-touch background image value from the input image.

The results are shown in \autoref{fig:confusion} and in Table~\ref{tab:material-accuracy-precision}. Our classifier achieves an overall accuracy of 72\%, with most errors coming from misalignment between acrylic and 3D print. We hypothesize that this is because the 3D print's texture in both real-life and simulation data is only faintly visible. Hence, the model often mistakes it for acrylic or wood. Additionally, the model misclassified several fabric textures as rock, potentially because the sampled objects had deformable surfaces, producing noisier features that approximate rock. Nonetheless, zero-shot classification achieves significantly higher accuracy than random guessing, supporting our claim that \normtotex\ can improve sim-to-real transfer.

\begin{figure}[t]
\vspace{5.1pt}
  \centering
    \includegraphics[width=0.95\linewidth]{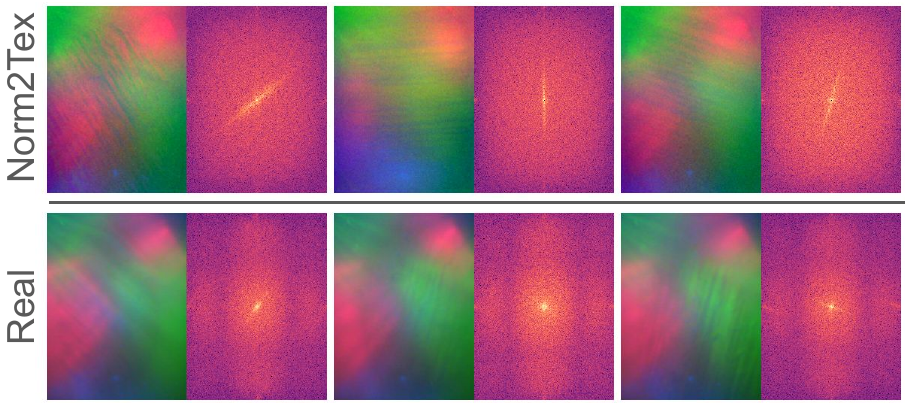}
    \caption{\normtotex (top) and real-world (bottom) tactile images and 2D Fourier transform for wood. \normtotex\ retains directional/high-frequency spectral structure corresponding to the wood grain, while the real tactile capture is dominated by low frequencies when the texture is faint.}
    \label{fig:fft}
\end{figure}

Although \normtotex\ introduces material-specific surface structure into simulated tactile observations, Fig.~\ref{fig:fft} shows that similar-looking patterns between real and sim data can still produce different features for the downstream pipeline.
In simulation, the wood texture produces pronounced directional and high-frequency components in the Fourier spectrum. Because the real world is inherently noisy, these components are attenuated in real DIGIT images, especially when the texture is faint.
This shows that the tested materials cannot fully reproduce real-world visuals, but strong classification performance suggests that perfect visual matching is not necessary for effective sim-to-real transfer.

\subsection{Zero-shot Sim-to-Real Robot Policy}

To demonstrate the importance of texture in tactile data for robotics, we conduct a closed-loop policy experiment. 
The proposed task is to hold a cube with the lightest possible force, i.e., the largest opening of the gripper, without letting the cube fall. The objective of this experiment is not to obtain an optimal release policy, but to evaluate whether simulated tactile texture leads to differences in closed-loop behavior after transfer.
This task requires the policy to identify the cube's state and determine whether slipping is likely if it opens one more step.
The friction coefficient depends on the material, which is easy to identify by texture.
The policy is formulated as a discrete classification problem between open, close, or maintain the gripper position. As the gripper's state is normalized to a range of 0 (fully closed) and 1 (fully open), this defines the policy's action space as a delta command of $+0.005$ for open, $-0.005$ for close, and $0.000$ for stay. 
The observation space contains the normalized gripper width and the left and right digit images.

\begin{figure}[t]
  \vspace{5.1pt}
  \centering
    \includegraphics[width=\linewidth]{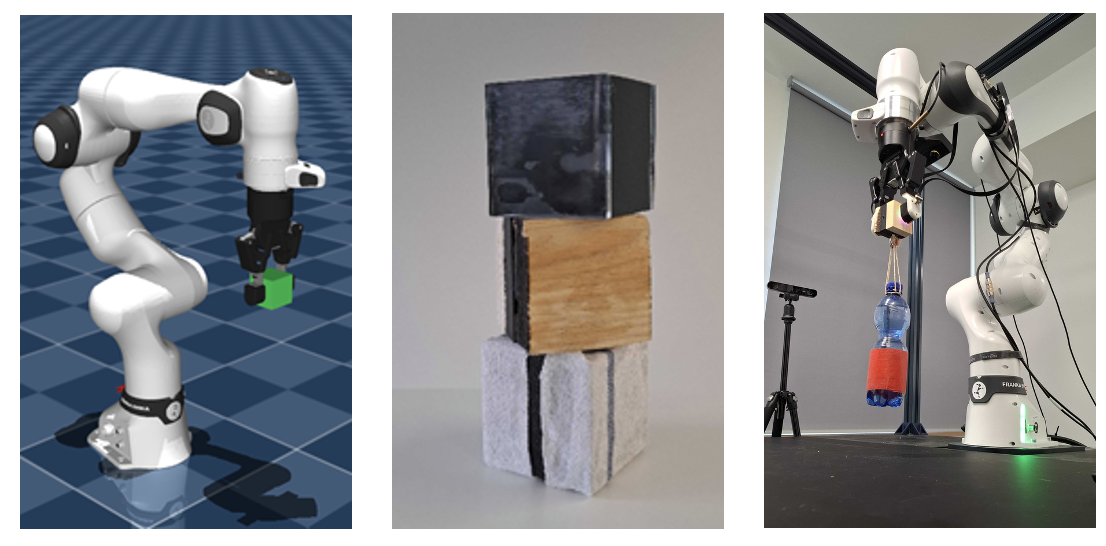}
        \caption{(Left) Simulation setup for the holding cube experiment inside MuJoCo simulation. (Center) Cubes with wood, acrylic, and fabric wrapping materials used for real-world grasping. (Right) Real-world robot setup. Note that neither the side nor the wrist camera is needed in the experiments.}
        \label{fig:setup}
\end{figure}

The test cube is composed of three of the most distinguishable materials already classified in the sim-to-real experiment: wood, fabric, and acrylic. 
In all experiments, the cubes have the same weight. 
In the real world, we attach a 500~ml bottle of water to each cube to induce slippage.
Fig.~\ref{fig:setup} shows the constructed cubes and the setup used for the real-world experiments.

\begin{figure}[t]
  \vspace{5.1pt}
  \centering
    \includegraphics[width=0.23\linewidth]{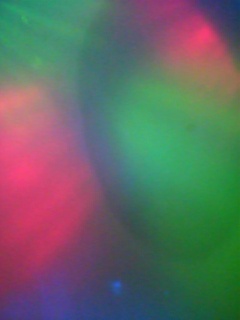}
    \includegraphics[width=0.23\linewidth]{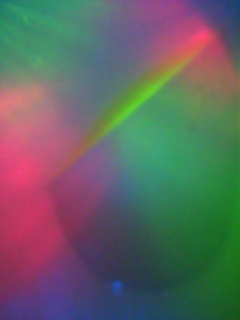}
    {\vrule width 1.5pt height 2.7cm}
    \includegraphics[width=0.23\linewidth]{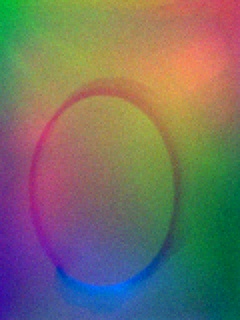}
    \includegraphics[width=0.23\linewidth]{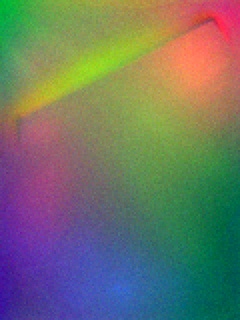}
        \caption{(Left) Real DIGIT images of the acryl box's surface and edge. (Right) Simulated textureless tactile images of an equal-sized box's surface and edge. Note that the shadow and lighting caused by the compression of the silicon pad are not equally reproduced on Taxim.}
        \label{fig:acryl}
\end{figure}

We train using PPO~\cite{schulman2017proximal,stable-baselines3} over a 128-step horizon.
After holding the cube with maximum pressure, we set the gripper command to a random value between 0.45 and 0.55.
To collect varied data, we randomize the initial relative height of the cube's center of mass to the digit center between 0 and 2.5cm.
We also rotate between 24 different variations of each texture.
The episode is terminated as soon as the cube drops out of the gripper.

We use a dense reward based on the gripper command. 
For non-terminal states, we compute the reward as a scaled sigmoid of the gripper opening, encouraging the policy to release the object progressively. 
The sigmoid function starts at 0.5, reaches its midpoint at 0.6, and saturates at 0.7.
Hence, each safe increment in gripper opening gradually increases the reward.
If the object is dropped, the episode terminates, and the reward is set to 0.

We ablate this policy in three variants: a blind policy that inputs only the gripper opening state; a policy with tactile but no texture, i.e., Taxim rendering; and a third policy with visual, tactile, and texture, i.e., \normtotex.
We background-subtracted each tactile image using a corresponding blank sensor image and processed it with the same ResNet-18 encoder.
We start with the pre-trained weights from the classifier task and fine-tune them using a learning rate ten times lower than that of the actor-critic feed-forward head.
We concatenated the two visual embeddings with the gripper observation encoding and passed them to the actor and critic. 

\begin{figure*}[t]
  \vspace{5.1pt}
  \centering
    \includegraphics[width=\linewidth]{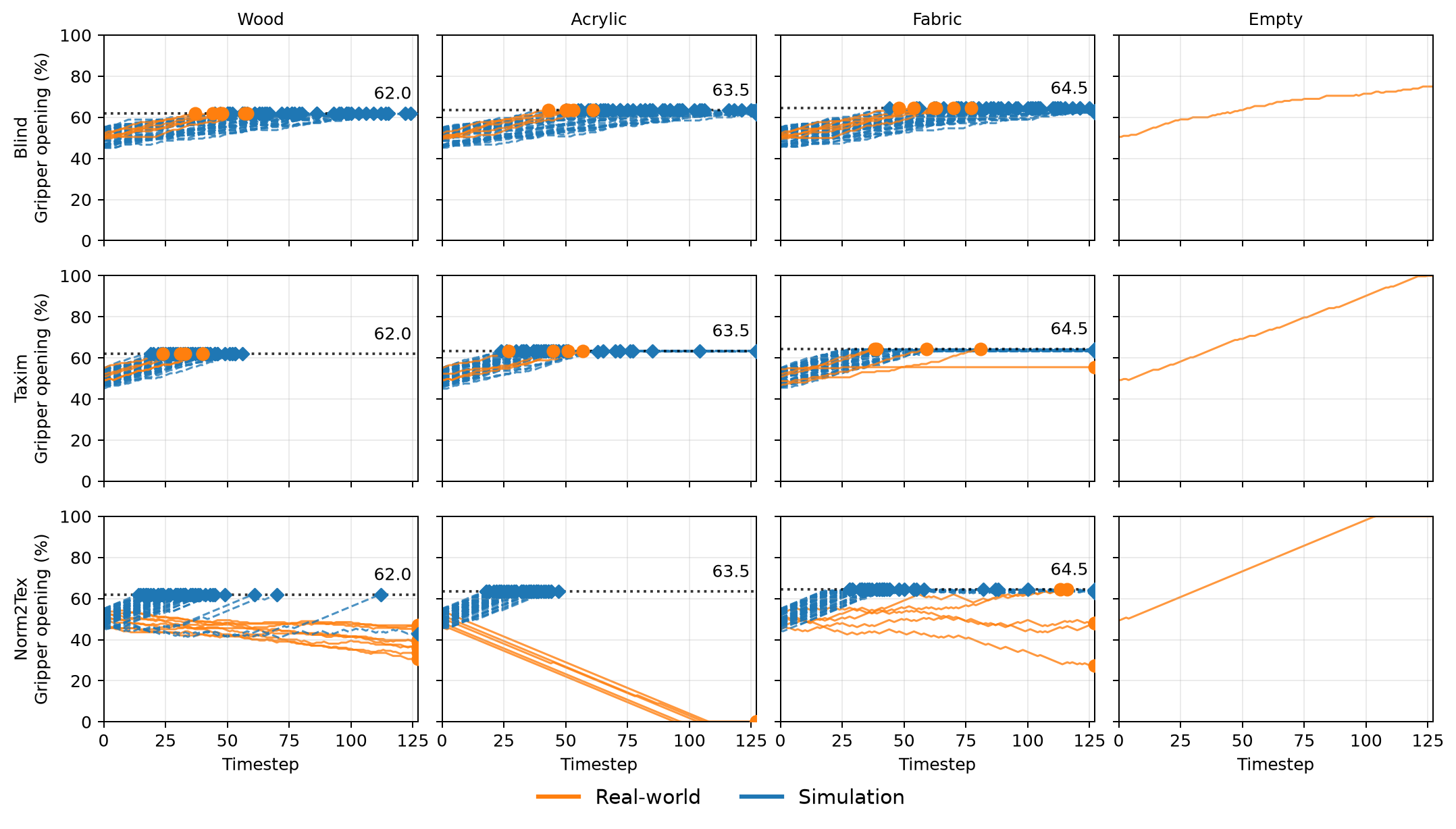}
        \caption{Gripper state for real-world and simulation rollouts of the cube-holding experiment for blind, Taxim, and \normtotex\ policies using wood, acrylic, and fabric materials. In the empty condition, all policies continue opening because no object-induced termination occurs. The blind and Taxim policies exhibit similar opening behavior across materials, whereas \normtotex\ produces distinct material-dependent responses in the real world. These results indicate that \normtotex\ observes tactile distinctions between grasp conditions. The black lines and values represent the maximum gripper opening before dropping. Those values were determined empirically.}
        \label{fig:results-rl}
\end{figure*}

Taxim and \normtotex\ policies used the same policy architecture. They differed only in the simulated tactile observations supplied to the network.
For each material, we empirically determined the maximum gripper command before dropping. 
The resulting gripper-command trajectories for simulated and real-world are shown in Fig.~\ref{fig:results-rl}. 
Among the evaluated materials, wood had the lowest release threshold, followed by acrylic and fabric.

The blind policy shows a consistent tendency to open in both real-world and simulated environments.
This outcome is expected because the policy only observes the gripper state and is unable to detect object release.
It is therefore driven by the reward gradient to continually increase gripper opening until release occurs.
Across all policies, the empty condition leads to opening. 
None of the policies learns to treat the absence of an object as a reason to stop.
This is an out-of-domain behavior, since there is no example in training where the cube is not being held besides in the termination state.

The Taxim-based policy also produces largely material-independent actions. 
In simulation, the policy learned to overfit to the fabric threshold, which is the highest among all materials.
This result suggests that the tactile signals were not informative enough for the policy to learn.
We hypothesize that this is due to the reward design, which encourages larger openings but does not penalize drops. 

In simulation, \normtotex\ sustains longer rollouts on wood and fabric and, unlike Taxim, does not converge to a single material-independent opening threshold.
Although none of the policies perfectly optimizes the task, the contrast between variants with and without texture is informative.
In the real world, the policy closes moderately on wood, closes strongly on acrylic, and maintains larger openings on fabric. 
While the real-world trajectories do not exactly reproduce those observed in simulation, the material-dependent behavior indicates that texture provides transferable tactile cues for closed-loop control. 
Qualitative sim-to-real agreement is strongest for wood and fabric, while the strong closing response on acrylic points to a remaining sim-to-real gap.

We hypothesize that this sim-to-real gap for acrylic originates from deformation-dependent illumination artifacts in the real DIGIT sensor. 
As shown in Fig. \ref{fig:acryl}, compression of the sensor pad generates shadow near the contact boundary.
Taxim does not model this effect.
Because the acrylic surface provides little texture, these shadows dominate the visual representation and shift the tactile features away from the simulation distribution that the model was trained on. 

\section{Conclusions}
In this work, we propose \normtotex, a plugin to add texture to visuo-tactile rendering.
\normtotex\ incorporates high-frequency surface details by modifying the normal map of the depth representation without affecting the original simulation pipeline.
We demonstrate sim-to-real transfer by achieving around 72\% accuracy on a zero-shot material classifier trained solely on simulation renderings.
We also assess the performance of the texture rendering in a robot policy that depends on material classification.
Compared with texture-less simulations, the reinforcement learning policy based on\normtotex\ can choose different behaviors depending on the grasped material.
Future work can focus on implementing this logic into FEM/MPM-based tactile simulators, and reproducing complex contact shadowing and deformation-dependent patterns to further close the gap between simulated and real visual-tactile sensing.
We will make our procedural generation pipeline for arbitrary meshes open-source after acceptance of this manuscript.

\section*{Acknowledgments}
\blackout{
 The authors gratefully acknowledge the scientific support and HPC resources provided by the Erlangen National High Performance Computing Center (NHR@FAU) of the Friedrich-Alexander-Universität Erlangen-Nürnberg (FAU) under the BayernKI project v106be. BayernKI funding is provided by Bavarian state authorities. This work has been partially supported by the German Federal Ministry of Research, Technology and Space (BMFTR) under the Robotics Institute Germany (RIG). This work has been partially supported by the project GeniusRobot funded by the German Federal Ministry of Education and Research (BMBF grant no. 01IS24083).} The authors acknowledge the use of ChatGPT for portions of code creation. All generated code was rigorously reviewed, adapted and validated by the authors. All methodologies, evaluations and interpretation of results were designed and performed by the authors.

\bibliographystyle{IEEEtran}
\bibliography{bibliography}

@IEEEtranBSTCTL{IEEEexample:BSTcontrol,
CTLuse_forced_etal = "yes",
CTLmax_names_forced_etal = "6",
CTLnames_show_etal = "5",
CTLdash_repeated_names = "no"
}

@string{ieeetro = "IEEE Transactions on Robotics"}

@string{ieeeral = "IEEE Robotics and Automation Letters"}

@string{IROS = "Proc.~of the IEEE/RSJ Int.~Conf.~on Intelligent Robots and Systems (IROS)"}

@string{ICRA = "Proc.~of the IEEE Int.~Conf.~on Robotics \& Automation (ICRA)"}

@string{CVPR = "Proc.~of the IEEE Computer Society Conference on
                  Computer Vision and Pattern Recognition (CVPR)"}

@string{jmlr = "Journal of Machine Learning Research"}

@string{ICLR = "Proc.~of the Int.~Conf.~on Learning Representations (ICLR)"}

@inproceedings{he2016deep,
  title={Deep residual learning for image recognition},
  author={He, Kaiming and Zhang, Xiangyu and Ren, Shaoqing and Sun, Jian},
  booktitle=CVPR,
  year={2016}
}

@article{schulman2017proximal,
  title={Proximal policy optimization algorithms},
  author={Schulman, John and Wolski, Filip and Dhariwal, Prafulla and Radford, Alec and Klimov, Oleg},
  journal={arXiv preprint arXiv:1707.06347},
  year={2017}
}

@Article{gelsight,
AUTHOR = {Yuan, Wenzhen and Dong, Siyuan and Adelson, Edward H.},
TITLE = {{GelSight}: High-Resolution Robot Tactile Sensors for Estimating Geometry and Force},
JOURNAL = {Sensors},
YEAR = {2017},
}

@inproceedings{improvedgelsight,
author = {Dong, Siyuan and Yuan, Wenzhen and Adelson, Edward H.},
title = {Improved GelSight tactile sensor for measuring geometry and slip},
year = {2017},
booktitle = IROS,
}

@ARTICLE{digit,
  author={Lambeta, Mike and Chou, Po-Wei and Tian, Stephen and Yang, Brian and Maloon, Benjamin and Most, Victoria Rose and Stroud, Dave and Santos, Raymond and Byagowi, Ahmad and Kammerer, Gregg and Jayaraman, Dinesh and Calandra, Roberto},
  journal=ieeeral, 
  title={{DIGIT}: A Novel Design for a Low-Cost Compact High-Resolution Tactile Sensor With Application to In-Hand Manipulation}, 
  year={2020},
}

@inproceedings{wang2021wedge,
  title={GelSight Wedge: Measuring High-Resolution 3D Contact Geometry with a Compact Robot Finger},
  author={Wang, Shaoxiong and She, Yu and Romero, Branden and Adelson, Edward H},
  booktitle=ICRA,
  year={2021},
}

@misc{liu2026crossmodalvisuotactilerepresentationlearning,
      title={Cross-Modal Visuo-Tactile Representation Learning with Action Chunking Transformers for Contact-Rich Manipulation}, 
      author={Yaohua Liu and Rong Fu and Amir H. Gandomi and Simon Fong and Hengjun Zhang},
      year={2026},
      journal={arXiv preprint arXiv:2602.00514},
}

@misc{higuera_perceiving_2023,
	title = {Perceiving {Extrinsic} {Contacts} from {Touch} {Improves} {Learning} {Insertion} {Policies}},
    journal={arXiv preprint arXiv:2309.16652},
	author = {Higuera, Carolina and Ortiz, Joseph and Qi, Haozhi and Pineda, Luis and Boots, Byron and Mukadam, Mustafa},
	year = {2023}
}

@INPROCEEDINGS{mujoco,
  author={Todorov, Emanuel and Erez, Tom and Tassa, Yuval},
  booktitle=IROS, 
  title={{MuJoCo}: A physics engine for model-based control}, 
  year={2012}
}

@ARTICLE{tacflex_2025,
  author={Zhang, Chaofan and Cui, Shaowei and Hu, Jingyi and Jiang, Tianyu and Zhang, Tiandong and Wang, Rui and Wang, Shuo},
  journal=ieeetro, 
  title={{TacFlex}: Multimode Tactile Imprints Simulation for Visuotactile Sensors With Coating Patterns}, 
  year={2025}
}

@ARTICLE{tacchi_2023,
  author={Chen, Zixi and Zhang, Shixin and Luo, Shan and Sun, Fuchun and Fang, Bin},
  journal=ieeeral, 
  title={Tacchi: A Pluggable and Low Computational Cost Elastomer Deformation Simulator for Optical Tactile Sensors}, 
  year={2023}
}

@ARTICLE{si_taxim_2021,
  author={Si, Zilin and Yuan, Wenzhen},
  journal=ieeeral, 
  title={Taxim: An Example-Based Simulation Model for GelSight Tactile Sensors}, 
  year={2022},
}

@ARTICLE{Wang2022TACTO,
  author={Wang, Shaoxiong and Lambeta, Mike and Chou, Po-Wei and Calandra, Roberto},
  journal=ieeeral, 
  title={{TACTO}: A Fast, Flexible, and Open-Source Simulator for High-Resolution Vision-Based Tactile Sensors}, 
  year={2022}
}

@ARTICLE{tacipc_2024,
  author={Du, Wenxin and Xu, Wenqiang and Ren, Jieji and Yu, Zhenjun and Lu, Cewu},
  journal=ieeeral, 
  title={TacIPC: Intersection- and Inversion-Free FEM-Based Elastomer Simulation for Optical Tactile Sensors}, 
  year={2024},
  volume={9},}

@article{tacchi2_2026,
title = {Tacchi 2.0: A low computational cost and comprehensive dynamic contact simulator for vision-based tactile sensors},
journal = {Sensors and Actuators},
year = {2026},
author = {Yuhao Sun and Shixin Zhang and Wenzhuang Li and Shaobo Yang and Derui Li and Jiatong Du and Di Guo and Jianwei Zhang and Bin Fang},}

@ARTICLE{tacsl_2025,
  author={Akinola, Iretiayo and Xu, Jie and Carius, Jan and Fox, Dieter and Narang, Yashraj},
  journal=ieeetro, 
  title={TacSL: A Library for Visuotactile Sensor Simulation and Learning}, 
  year={2025},}

@inproceedings{difftactile_2024,
  author       = {Zilin Si and
                  Gu Zhang and
                  Qingwei Ben and
                  Branden Romero and
                  Zhou Xian and
                  Chao Liu and
                  Chuang Gan},
  title        = {{DIFFTACTILE:} A Physics-based Differentiable Tactile Simulator for Contact-rich Robotic Manipulation},
  booktitle    =  ICLR,
  year         = {2024}
}

@ARTICLE{texture_cyclegan_2022,
  author={Chen, Weihang and Xu, Yuan and Chen, Zhenyang and Zeng, Peiyu and Dang, Renjun and Chen, Rui and Xu, Jing},
  journal=ieeeral, 
  title={Bidirectional Sim-to-Real Transfer for GelSight Tactile Sensors With CycleGAN}, 
  year={2022},}

@inproceedings{jianu_reducing_2022,
	title = {Reducing {Tactile} {Sim2Real} {Domain} {Gaps} via {Deep} {Texture} {Generation} {Networks}},
	booktitle = ICRA,
	author = {Jianu, Tudor and Gomes, Daniel Fernandes and Luo, Shan},
	year = {2022},
}

@article{chen_bidirectional_2022,
	title = {Bidirectional {Sim}-to-{Real} {Transfer} for {GelSight} {Tactile} {Sensors} {With} {CycleGAN}},
	journal = ieeeral,
	author = {Chen, Weihang and Xu, Yuan and Chen, Zhenyang and Zeng, Peiyu and Dang, Renjun and Chen, Rui and Xu, Jing},
	year = {2022},
}

@inproceedings{azulay_augmenting_2024,
	title = {Augmenting {Tactile} {Simulators} with {Real}-like and {Zero}-{Shot} {Capabilities}},
	booktitle = ICRA,
	author = {Azulay, Osher and Mizrahi, Alon and Curtis, Nimrod and Sintov, Avishai},
	year = {2024},
}

@ARTICLE{lin_diffusion_2025,
  author={Lin, Xi and Xu, Weiliang and Mao, Yixian and Wang, Jing and Lv, Meixuan and Liu, Lu and Luo, Xihui and Li, Xinming},
  journal={IEEE Transactions on Instrumentation and Measurement}, 
  title={Vision-Based Tactile Image Generation via Contact Condition-Guided Diffusion Model}, 
  year={2025},}

@article{frankot1988,
author = {Frankot, Robert and Chellappa, Rama},
year = {1988},
title = {A method for enforcing integrability in shape from shading algorithms},
journal = {IEEE Transactions on Pattern Analysis and Machine Intelligence},
}

@inproceedings{CycleGAN2017,
  title={Unpaired Image-to-Image Translation using Cycle-Consistent Adversarial Networkss},
  author={Zhu, Jun-Yan and Park, Taesung and Isola, Phillip and Efros, Alexei A},
  booktitle=ICCV,
  year={2017}
}

@article{Cohen1998AppearancepreservingS,
  title={Appearance-preserving simplification},
  author={Jonathan D. Cohen and Marc Olano and Dinesh Manocha},
  journal={Proceedings of the 25th annual conference on Computer graphics and interactive techniques},
  year={1998},
}

@software{blender,
  title = {Blender: A 3D modelling and rendering package},
  organization = {Blender Foundation},
  year = {2026},
  version = {5.1.2}
}

@ARTICLE{fabricsorting2026,
  author={Li, Jiayao and Gao, Yu and Yan, Yijia and Li, Zhenke and Wu, Xin and Huang, Jipeng},
  journal={IEEE Sensors Journal}, 
  title={Vision–Tactile Sensor Fusion System for Fabric Sorting and Robotic Grasping in Textile Recycling}, 
  year={2026},}

@article{defectdetection2026,
title = {Tac-TSS-OCD: Tactile-based surface micron defect detection in open environments},
journal = {Advanced Engineering Informatics},
year = {2026},
author = {Zerui Xi and Yiping Gao and Liang Gao and Xinyu Li and Boyang Hu},
}

@ARTICLE{fabricdefect2022,
  author={Fang, Bin and Long, Xingming and Sun, Fuchun and Liu, Huaping and Zhang, Shixin and Fang, Cheng},
  journal={IEEE Transactions on Instrumentation and Measurement}, 
  title={Tactile-Based Fabric Defect Detection Using Convolutional Neural Network With Attention Mechanism}, 
  year={2022},}

@inproceedings{whatmatters2024,
    author = {Boehm, Alina and Schneider, Tim and Belousov, Boris and Kshirsagar, Alap and Lin, Lisa and Doerschner, Katja and Drewing, Knut and Rothkopf, Constantin A. and Peters, Jan},
    title = {What Matters for Active Texture Recognition With Vision-Based Tactile Sensors},
    booktitle = ICRA,
    year = {2024},
}

@article{stable-baselines3,
  author  = {Antonin Raffin and Ashley Hill and Adam Gleave and Anssi Kanervisto and Maximilian Ernestus and Noah Dormann},
  title   = {Stable-Baselines3: Reliable Reinforcement Learning Implementations},
  journal = jmlr,
  year    = {2021},
}

\end{document}